\PassOptionsToPackage{dvipsnames,table}{xcolor}
\PassOptionsToPackage{capitalise}{cleveref}
\PassOptionsToPackage{expansion=false}{microtype}
\documentclass{spml_template}
\pdfoutput=1

\usepackage{amsmath}
\usepackage{amssymb}
\usepackage{mathtools}
\usepackage{amsthm}

\usepackage{amsmath,amsfonts,bm}

\def\eqref#1{equation~\ref{#1}}

\def\1{\bm{1}}

\DeclareMathAlphabet{\mathsfit}{\encodingdefault}{\sfdefault}{m}{sl}
\SetMathAlphabet{\mathsfit}{bold}{\encodingdefault}{\sfdefault}{bx}{n}

\DeclareMathOperator{\softmax}{softmax}

\DeclareMathOperator*{\argmax}{arg\,max}

\usepackage{enumitem}
\usepackage{algorithm}
\usepackage{algorithmic}
\usepackage{url}
\crefname{appendix}{Appendix}{Appendices}
\Crefname{appendix}{Appendix}{Appendices}

\theoremstyle{plain}

\theoremstyle{definition}

\newcommand{\sys}{HACO}
\newcommand{\syslong}{Human--AI Co-discovery system}
\newcommand{\sysfull}{\sys{} (\syslong)}

\usepackage{tikz}
\usetikzlibrary{shapes.geometric}
\usepackage[most]{tcolorbox}
\definecolor{agentteal}{RGB}{34,128,118}
\definecolor{agentfill}{RGB}{240,247,246}
\definecolor{humanorange}{RGB}{190,108,32}
\definecolor{humanfill}{RGB}{252,246,238}
\definecolor{orchslate}{RGB}{85,102,122}
\definecolor{orchfill}{RGB}{242,243,245}
\definecolor{nodegray}{RGB}{120,120,120}
\newcommand{\roboticon}[1]{%
\begin{tikzpicture}[scale=0.16,line width=0.9pt]
  \draw[#1,rounded corners=1pt,fill=#1!12] (-1.6,-1.6) rectangle (1.6,1.4);
  \draw[#1] (0,1.4)--(0,2.1); \fill[#1] (0,2.25) circle (0.28);
  \fill[#1] (-0.7,0.1) circle (0.42); \fill[#1] (0.7,0.1) circle (0.42);
  \draw[#1,line width=0.7pt] (-0.7,-0.9)--(0.7,-0.9);
\end{tikzpicture}}
\newcommand{\humanicon}[1]{%
\begin{tikzpicture}[scale=0.16,line width=0.9pt]
  \fill[#1] (0,0.55) circle (0.62);
  \draw[#1,fill=#1] (-1.25,-1.7) .. controls (-1.25,-0.35) and (1.25,-0.35) .. (1.25,-1.7) -- cycle;
\end{tikzpicture}}
\newcommand{\opchip}[2]{\tikz[baseline=-0.5ex]{\node[fill=#1!14,draw=#1,line width=0.5pt,rounded corners=2pt,inner xsep=4pt,inner ysep=1.5pt,font=\footnotesize\bfseries,text=#1]{#2};}}
\newcommand{\nodetag}[1]{{\footnotesize\color{nodegray}(#1)}}
\newtcolorbox{agentturn}{enhanced,breakable,left=10mm,top=1.8mm,bottom=1.8mm,right=2.5mm,
  colback=agentfill,colframe=agentteal,boxrule=0.6pt,arc=2.5pt,
  overlay={\node[anchor=north west] at ([xshift=2mm,yshift=-1.8mm]frame.north west){\roboticon{agentteal}};}}
\newtcolorbox{humanturn}{enhanced,breakable,left=10mm,top=1.8mm,bottom=1.8mm,right=2.5mm,
  colback=humanfill,colframe=humanorange,boxrule=0.6pt,arc=2.5pt,
  overlay={\node[anchor=north west] at ([xshift=2mm,yshift=-1.8mm]frame.north west){\humanicon{humanorange}};}}
\newtcolorbox{orchturn}{enhanced,breakable,left=10mm,top=1.8mm,bottom=1.8mm,right=2.5mm,
  colback=orchfill,colframe=orchslate,boxrule=0.6pt,arc=2.5pt,
  overlay={\node[anchor=north west] at ([xshift=2mm,yshift=-1.8mm]frame.north west){\roboticon{orchslate}};}}
\newcommand{\agenthead}[1]{{\bfseries\color{agentteal}Agent}\;\opchip{agentteal}{#1}\par\smallskip}
\newcommand{\humanhead}{{\bfseries\color{humanorange}Human}\par\smallskip}
\newcommand{\orchhead}{{\bfseries\color{orchslate}Orchestrator}\par\smallskip}

\title{Discovering Crystal Structure Prediction \\
 Algorithms with an AI Co-Scientist}

\author[1]{Kiyoung Seong}
\author[1]{Nayoung Kim}
\author[1]{Sungsoo Ahn}

\affiliation[1]{Korea Advanced Institute of Science and Technology (KAIST)}

\metadata[Code]{\href{https://github.com/kiyoung98/HACO}{https://github.com/kiyoung98/HACO}, \href{https://github.com/kiyoung98/maskgxt}{https://github.com/kiyoung98/maskgxt}}
\metadata[Blog]{\href{https://sungsoo-ahn.github.io/blog/2026/maskgxt-ai-coscientist/}{https://sungsoo-ahn.github.io/blog/2026/maskgxt-ai-coscientist/}}
\correspondence{\email{kiyoung.seong@kaist.ac.kr}, \email{sungsoo.ahn@kaist.ac.kr}}

\abstract{
We introduce \syslong{} (\sys{}) for scientific algorithm discovery through cross-domain search and sparse human steering. Starting from the goal of generating crystal structures from chemical compositions, \sys{} searched across generative modeling methodologies from multiple fields and identified MaskGIT, a masked generative model from vision, as a promising framework for crystal structure prediction (CSP). \sys{} instantiated this masked formulation as a discrete token model of crystal structure; guided by sparse high-level human objectives, it then added crystallographic symmetry tokens, space group stratified sampling for polymorph coverage, and sub-bin coordinate refinement, yielding the Masked Generative Crystal Transformer (MaskGXT). On the MP-20 polymorph split, MaskGXT reaches 79.06\% match-everyone-to-reference (METRe) accuracy, compared with 70.87\% for the strongest evaluated baseline. MaskGXT also attains the best match rate on standard MP-20 and MPTS-52 CSP benchmarks. These results provide evidence that, in domains offering cheap, fast, and well-aligned validation, transfer-guided interactive AI co-scientists can contribute to scientific algorithm discovery by identifying transferable modeling principles and combining them with targeted human domain guidance.
}

\newcommand{\pscrmseomatg}{0.205}
\newcommand{\pscrmsecryst}{0.198}
\newcommand{\pscrmsemask}{0.158}

\begin{document}

\maketitle

\section{Introduction}
\label{sec:introduction}

Artificial intelligence is beginning to act not only as a tool for scientific modeling, but also as a partner in scientific algorithm discovery~\citep{lu2024aiscientist,yamada2025aiscientistv2}. Recent systems can read literature, propose hypotheses, write code, and iterate on empirical feedback, producing progress in mathematics~\citep{romeraparedes2024funsearch,novikov2025alphaevolve}, biomedicine~\citep{gottweis2025coscientist}, and data science~\citep{grosnit2024agentk}. A distinctive advantage of such systems is breadth: modern AI models are trained on knowledge spanning computer vision, language, programming, and the sciences, and can therefore suggest transfers across fields that a domain specialist may not naturally consider.

A central open question is whether this breadth can produce a state-of-the-art algorithm on a realistic and competitive scientific machine-learning benchmark. Existing ML engineering agents primarily operate within a fixed problem specification: the benchmark, input--output interface, evaluation metric, and often the broad modeling family are given, and the agent searches for better code, hyperparameters, or implementation variants~\citep{wijk2024rebench,jiang2025aide,toledo2025airesearchagents,chan2025mlebench,nathani2025mlgym}. Existing AI-scientist systems broaden this loop to hypothesis generation and paper drafting, but their strongest demonstrations are often evaluated as autonomous research workflows rather than as new algorithms that improve the state of the art in a mature scientific domain. We target a stricter setting: can an AI co-scientist, interacting with sparse human domain guidance, identify a new modeling principle, adapt it to the constraints of a scientific problem, and produce a competitive algorithm that advances standard domain benchmarks?

To study this question, we introduce \sysfull{} for scientific algorithm discovery. \sys{} is built around two principles. First, it performs cross-domain transfer: rather than only perturbing an existing domain model, it searches across generative modeling frameworks from other fields and evaluates whether they can be adapted to the target scientific problem. Second, it supports sparse human interaction: the human supplies high-level mechanisms or objectives, such as domain constraints or desired failure-mode corrections, while the agent implements, trains, evaluates, and refines candidate algorithms. 

We instantiate \sys{} in crystal structure prediction (CSP). CSP asks for a stable crystal structure given a chemical composition, and is a central problem in computational materials discovery~\citep{butler2018machine, oganov2019structure}. It is also a competitive setting within modern crystal generative modeling: recent progress includes early periodic material generators~\citep{xie2021crystal}, CSP-specific diffusion models~\citep{jiao2023crystal}, Riemannian flow and stochastic-interpolant frameworks that benchmark both CSP and de novo generation~\citep{miller2024flowmm,hoellmer2025omat}, unified multimodal crystal generators covering CSP among other tasks~\citep{seong2026multimodal}, and symmetry-aware or Wyckoff-based approaches that impose crystallographic constraints during generation~\citep{jiao2024space,kazeev2025wyckofftransformer}. The problem is scientifically constrained: crystals are periodic, fractional coordinates live on a torus, space group symmetry affects validity and sample quality, and many compositions admit multiple polymorphs. Thus, improving CSP is not merely a code-optimization exercise; it requires choosing a generative formulation that respects geometry, periodicity, symmetry, and polymorph diversity.

\begin{figure}[t]
\centering
\includegraphics[width=\textwidth]{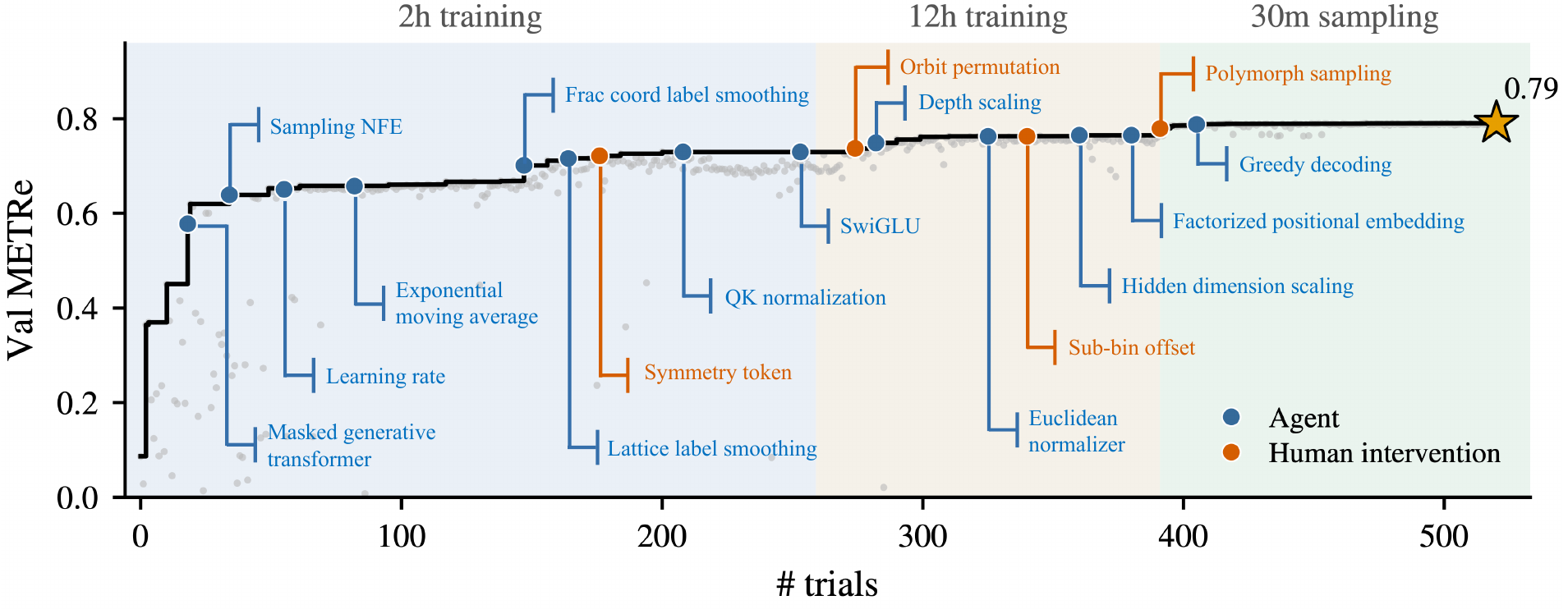}
\caption{\textbf{Research trajectory towards MaskGXT.}
The plot traces validation METRe against the number of trials. Each gray point is one candidate the agent trained and scored, and the black step line is the running best. Shaded regions
give the per-node budget, escalating from 2h to 12h training and then 30m sampling.}
\label{fig:search}
\end{figure}

Starting from the CSP task, \sys{} searched across generative modeling methodologies from multiple fields and identified MaskGIT~\citep{chang2022maskgit}, a masked generative model from vision, as a promising framework for crystals. Through interactive human--AI search, the agent adapted this framework into \emph{MaskGXT (Masked Generative Crystal Transformer)}\footnote{The ``X'' denotes the crystal, after \emph{Xtal}, the conventional shorthand for crystal.} (\cref{fig:search}). Human input was sparse and high-level, supplying crystallographic mechanisms or modeling objectives at a few points, such as symmetry-aware generation, polymorph coverage, and sub-bin coordinate precision, rather than low-level implementation.

MaskGXT casts CSP as masked Cdiscrete token generation. It represents lattices, fractional coordinates, space groups, and Wyckoff positions as tokens, and predicts them by masked parallel decoding. To make masked generative modeling suitable for crystals, MaskGXT combines periodicity-aware coordinate discretization, ordinal circular label smoothing, tokenized crystallographic symmetry, and sub-bin coordinate refinement. This discrete formulation enables confident greedy decoding and space group stratified sampling for polymorph coverage, providing an alternative to continuous score- or flow-matching generative models.

Empirically, MaskGXT attains state-of-the-art match rate on MP-20 and MPTS-52. On the MP-20 polymorph split, it reaches $79.06\%$ match-everyone-to-reference (METRe) accuracy, compared with $70.87\%$ for the strongest evaluated baseline. These results provide evidence that, in domains with cheap, fast, and well-aligned validation, an AI co-scientist paired with targeted human domain guidance can do more than automate implementation: it can help discover a competitive scientific algorithm by importing methodology across fields.

Our contributions are as follows:
\begin{itemize}[leftmargin=*]
\item We introduce \sys{} and provide evidence that such an AI co-scientist system can help discover a state-of-the-art algorithm for a realistic scientific machine-learning problem by transferring methodology from another field to crystal structure prediction.
\item We introduce MaskGXT, a masked generative crystal transformer discovered through this human--AI process, which represents lattices, coordinates, space groups, and Wyckoff positions as discrete tokens and adapts masked generative modeling to periodic, symmetry-constrained crystals.
\item We show that a single MaskGXT model attains state-of-the-art match rate on MP-20 and MPTS-52, and substantially improves METRe accuracy on the MP-20 polymorph split.
\end{itemize}

\section{Related Work}
\label{sec:related_work}

\paragraph{Crystal structure prediction models.}
Learning-based crystal structure prediction (CSP) has largely followed two modeling routes. The first treats lattice parameters and fractional coordinates as continuous variables and learns a generative process over periodic structures. Early periodic material generators such as CDVAE~\citep{xie2021crystal} introduced diffusion-based generation for crystals, and subsequent methods improved the geometric formulation through equivariant diffusion, Riemannian flow matching, stochastic interpolants, and related continuous generative processes~\citep{jiao2023crystal,miller2024flowmm,zeni2023mattergen,luo2025crystalflow,hoellmer2025omat}. These methods are natural for continuous geometry, but their sampling procedures typically require stochastic or iterative integration and do not directly expose a finite discrete decision space for greedy decoding.

The second route incorporates crystallographic symmetry more explicitly. Symmetry-aware methods generate or condition on space groups, Wyckoff positions, or symmetry templates~\citep{jiao2024space,zhu2024wycryst,crystalformer2025,kelvinius2025wyckoffdiff,kazeev2025wyckofftransformer}. These approaches place atoms in the unit cell under structural constraints, typically requiring post-processing and specialized architectures. MaskGXT differs in formulation: it represents the full CSP output, including lattice, coordinates, space group, and Wyckoff positions, as a single discrete token sequence decoded by a masked generative transformer. This lets symmetry information participate directly in masked token prediction, without a dedicated symmetry module, and enables space group-stratified decoding for multi-candidate generation.

\paragraph{Discrete and masked generative modeling.}
Discrete generative models corrupt a token sequence and learn to reverse the corruption. Discrete-diffusion models such as D3PM denoise a categorical process over many steps~\citep{austin2021d3pm}, while masked generative models corrupt by masking and reconstruct all positions in a few parallel steps, driving strong results in language and vision~\citep{devlin2019bert,chang2022maskgit}. Discrete representations have also entered crystal generation, but they discretize symmetry rather than geometry: WyckoffDiff runs categorical diffusion over space group and Wyckoff tokens without generating continuous coordinates~\citep{kelvinius2025wyckoffdiff}, while SymmCD discretizes only the site symmetry and atom type and keeps the coordinates continuous through diffusion~\citep{levy2025symmcd}. MaskGXT instead discretizes the coordinates themselves and decodes the full crystal, including coordinates, by confidence-ranked masked parallel prediction.

\paragraph{AI co-scientists and autonomous research agents.}
A growing body of work studies language-model agents that automate parts of the scientific research loop~\citep{wei2025agentic}. The AI Scientist and its successor generate ideas, write code, run experiments, analyze results, and draft papers with limited human intervention~\citep{lu2024aiscientist,yamada2025aiscientistv2}. Related systems emphasize different parts of the same loop: FunSearch and AlphaEvolve use program search to discover algorithms~\citep{romeraparedes2024funsearch,novikov2025alphaevolve}; AutoResearch repeatedly modifies training code, runs short experiments, and retains empirically improved variants~\citep{karpathy2026autoresearch}; and recent multi-agent systems such as AutoResearchClaw and AutoScientists explore collaboration, memory, debate, and long-horizon research trajectories~\citep{liu2026autoresearchclaw,gao2026autoscientists}. In parallel, AI engineering agentic benchmarks evaluate whether language-model agents can solve software- and machine-learning-development tasks by editing code, running experiments, and searching over implementation variants~\citep{wijk2024rebench,grosnit2024agentk,yang2024sweagent,jiang2025aide,toledo2025airesearchagents,chan2025mlebench,nathani2025mlgym}. 

Notably, our work differs from typical AI co-scientist systems in two ways. First, we use the agent to search for a transferable generative modeling principle for a scientific domain, instead of encouraging the agent to propose an entirely novel idea. Second, the process is interactive rather than fully autonomous: humans can hint at domain mechanisms or objectives for algorithm development. MaskGXT therefore serves as a case study in human--AI scientific algorithm discovery.

\section{\sys{}: \syslong{}}
\label{sec:agent}

\begin{figure}[t]
\centering
\includegraphics[width=\textwidth]{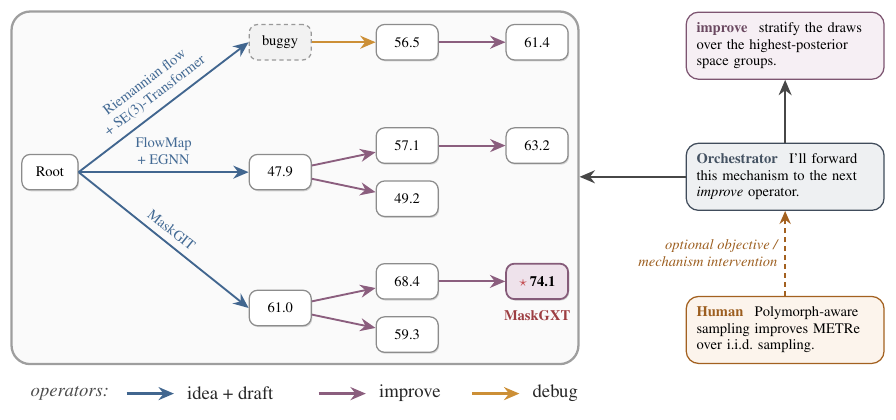}
\caption{\textbf{\sys{} for crystal structure prediction.} A single
orchestrator searches a tree of candidate CSP methods; each node is a candidate
generative model and the colored edges are search operators. The best lineage
transfers MaskGIT from vision, producing MaskGXT. A human sparsely intervenes
with high-level mechanisms or objectives, such as improving polymorph coverage.
Scores are illustrative validation METRe; full transcripts are provided in
\cref{app:dialogue}.}
\label{fig:agent}
\end{figure}

We first describe \sys{}, the AI co-scientist system that discovered MaskGXT (\cref{fig:agent}): the search task and objective, the tree-structured search, the operators, and where human guidance enters the loop.

\subsection{Task, Search, and Protocol}
\label{subsec:agent:task}

\paragraph{Task.}
The agent is asked to develop a model that generates crystal structures from chemical compositions. Its objective is to maximize validation METRe~\citep{martirossyan2025metre} on the MP-20 polymorph split, which rewards recovery of any reference polymorph for a given composition and so captures both structural accuracy and polymorph coverage. The entire development loop is confined to the MP-20 polymorph split: model proposals, component choices, and sampler refinements are selected only from its validation feedback.

\paragraph{Search.}
The core idea is to let an AI agent, rather than a human, decide \emph{which generative framework to use} for CSP. We build on recent AI engineering agents~\citep{jiang2025aide,toledo2025airesearchagents}, which treat method development as empirical search over code: the agent proposes a methodology, instantiates it as a CSP model, trains it, evaluates it against validation METRe, and revises the method. The search object is therefore not merely an implementation, but a generative modeling framework for CSP.

\paragraph{Protocol.}
The search proceeds in three stages of decreasing scope, moving from the methodology down to the sampler: each stage fixes what the previous one found and optimizes a finer level (\cref{fig:search}). The \emph{discovery} stage searches over the generative methodology, training each candidate under a short 2-hour budget while the full operator set runs until validation METRe stops improving; this stage selects the masked generative methodology. The \emph{scale-up} stage fixes this methodology and optimizes the model under a larger 12-hour per-candidate budget. The \emph{sampling optimization} stage fixes the model and tunes the sampling algorithm.

\subsection{Tree-Based Search and Orchestration}
\label{subsec:agent:tree}

\paragraph{Search tree.}
Each node in the search tree is a self-contained candidate: a complete training program that defines a generative model for CSP. The agent evaluates a node by training it under the stage-specific per-candidate budget and measuring validation METRe. It organizes nodes into a tree whose edges are search operators, expanding promising nodes and pruning unproductive ones so that strong lineages receive more compute over time. This mirrors the population- or tree-based search used by AI engineering agents~\citep{lu2024aiscientist,jiang2025aide,toledo2025airesearchagents}, with the empirical metric as the primary basis for selection.

\paragraph{Orchestrator.}
A single orchestrator, driven by Claude Opus 4.7~\citep{anthropic2026claudeopus47}, controls the search and serves as the sole interface for human steering. It sets up an isolated environment for each node, runs up to eight candidate nodes in parallel, one per GPU across eight RTX 3090 GPUs, decides which node to expand and which operator to apply; dispatches the operator; screens and evaluates the resulting candidate against validation METRe; and adds the candidate to the search tree. The loop runs autonomously, and when the human intervenes, the request enters through the orchestrator, which incorporates the message into the operator it dispatches (\cref{fig:agent}).

\subsection{Search Operators and Cross-Domain Transfer}
\label{subsec:agent:operators}

\paragraph{Operators.}
The agent expands the tree with four main operators. The \emph{idea} operator proposes a new methodology; \emph{draft} instantiates that methodology into runnable code; \emph{improve} makes one coherent change to an existing candidate; and \emph{debug} repairs a candidate that failed. Before training, \emph{draft} and \emph{improve} each predict why their code should raise validation METRe. Two auxiliary operators support the search. The \emph{smoke} operator runs a lightweight sanity check, using a few training and sampling steps on one batch. The \emph{analyze} operator interprets a finished run against that prediction, judges whether the result is valid, and records a finding that guides what to change next. All operators use the same Opus 4.7 as the orchestrator.

\paragraph{Cross-domain idea generation.}
The distinctive component is the \emph{idea} operator, which performs deliberate cross-domain transfer. Rather than perturbing the current method, it surveys the broader generative-modeling literature, across vision, language, and other fields, and returns (generative framework, architecture) pairs that have not been applied to crystals but have a credible mechanism for competing on CSP. The operator selects each pair under two criteria: novelty to CSP and adaptability to periodic crystals. In our run, it identified MaskGIT, a masked generative model from the vision domain~\citep{chang2022maskgit}, which the agent then instantiated and refined into a model for crystals, described next.

\subsection{Human Steering and Autonomy Boundary}
\label{subsec:agent:human}

The agent ran the full loop and wrote all model code itself, while the human intervened sparsely and only at a high level (\cref{fig:agent}). We categorize the human input by what it supplies. A \emph{mechanism} supplies domain knowledge the agent lacks~\citep{zou2025humanagent}, such as that crystal symmetry should be reflected or that non-i.i.d.\ sampling improves polymorph coverage. An \emph{objective} states a goal and leaves the agent to find the method~\citep{kwon2023reward}, such as recovering sub-bin coordinate precision. In both cases, the human supplies a high-level mechanism or goal, and the agent realizes it in code, including the domain adaptation. This keeps the autonomy claim calibrated: the agent performs the implementation, empirical search, and iterative refinement, while sparse human input supplies missing domain mechanisms or objectives.
\section{MaskGXT: Masked Generative Crystal Transformer}
\label{sec:maskgxt}

We now introduce MaskGXT, the crystal structure prediction model developed by AI co-scientist.
Rather than representing a crystal as continuous coordinates evolved through a diffusion or flow process, MaskGXT represents it as a discrete, partially masked sequence and learns to reconstruct the missing tokens.

\begin{figure}[t]
  \centering
  \includegraphics[width=\textwidth]{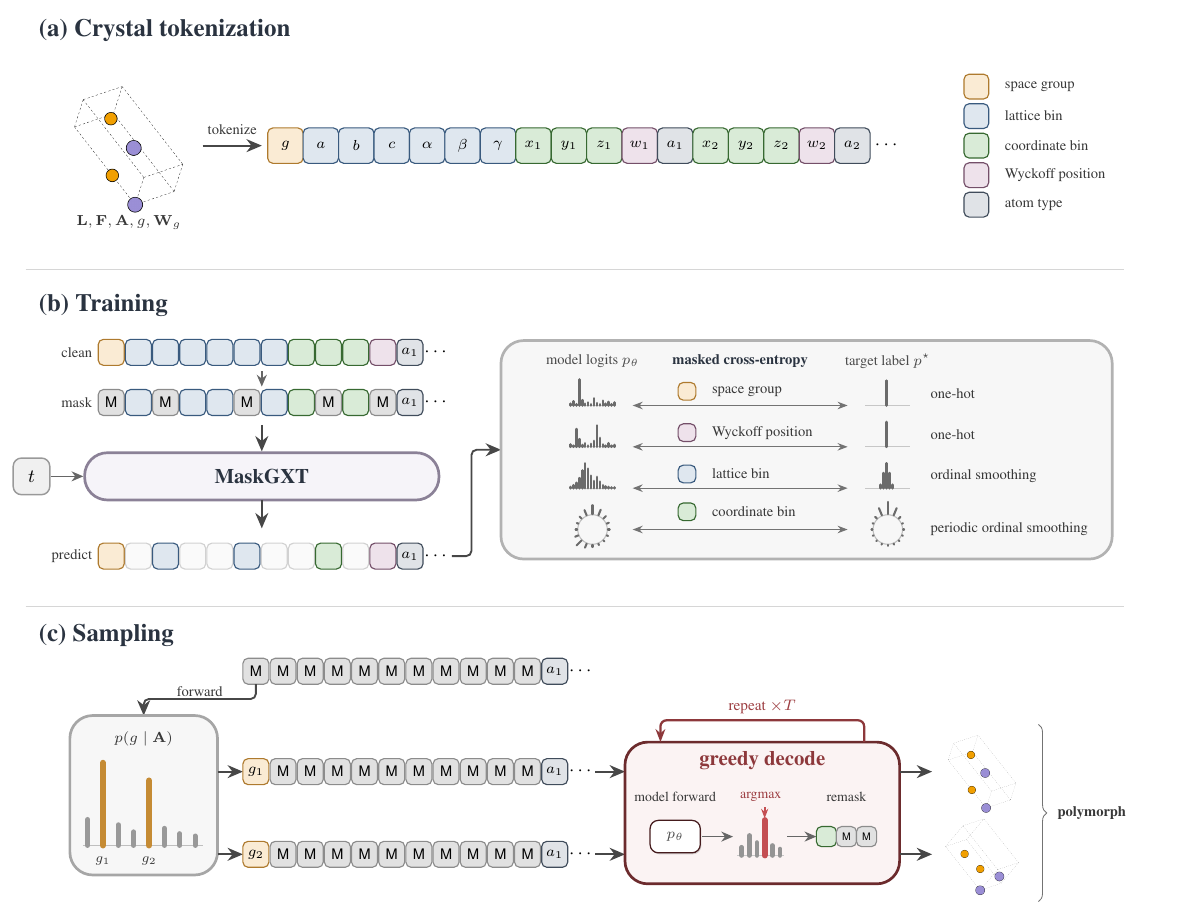}
  \caption{\textbf{MaskGXT tokenization, training, and sampling.}
  (a)~A crystal $(\mathbf{L},\mathbf{F},\mathbf{A},g,\mathbf{W}_g)$ is tokenized
  into $1{+}6{+}5N$ tokens: one space group token, six lattice tokens (lengths and
  angles), and, per site, three coordinate tokens, one Wyckoff token, and one
  atom-type token, with atom types $\mathbf{A}$ given as side information.
  (b)~A random fraction of tokens is masked and the transformer reconstructs
  them, trained with a masked cross-entropy loss.
  (c)~From a fully masked sequence with only $\mathbf{A}$ known, the space group
  posterior seeds distinct polymorph branches, and greedy decoding then unmasks
  the remaining tokens over $T$ steps.}
  \label{fig:maskgxt}
\end{figure}

However, directly applying MaskGIT to crystals is nontrivial, since crystals are periodic and symmetry-redundant. MaskGXT meets these challenges (\cref{fig:maskgxt}) with a discrete coordinate representation on the fractional-coordinate torus, symmetry-consistent augmentations, and confidence-ranked greedy decoding with space group stratified sampling. We detail each in turn: tokenization and discretization, symmetry-consistent descriptions, training objective, architecture, and decoding.

\subsection{Discrete Crystal Tokenization}
\label{sec:maskgxt_representation}

\paragraph{Crystal tokenization.}
In CSP, the input is a set of atom types $\mathbf{A}=(a_1,\dots,a_N)$, and the goal is to predict the lattice $\mathbf{L}$ and fractional coordinates $\mathbf{F}=(\mathbf{f}_1,\dots,\mathbf{f}_N)\in[0,1)^{3\times N}$. MaskGXT tokenizes a crystal into $1{+}6{+}5N$ tokens: one space group token, six lattice tokens for the quantized lengths and angles, and five tokens per site, namely three quantized coordinate tokens, a Wyckoff token, and an atom-type token (\cref{fig:maskgxt}). Atom types $\mathbf{A}$ are given as a condition, leaving the space group $g$, lattice, coordinate, and Wyckoff tokens $\mathbf{W}_g$ to predict. Symmetry is thus predicted jointly with the structure, rather than imposed through a separate module. Because the symmetry, lattice, and coordinate tokens are decoded independently, this consistency is encouraged through training rather than enforced as a hard constraint.

\paragraph{Coordinate discretization.}
To model crystal geometry with a masked transformer, MaskGXT represents continuous quantities as discrete tokens. It maps a fractional coordinate $f\in[0,1)$ to a discrete bin
\begin{equation}
q(f) = \left\lfloor (f \bmod 1)\,K \right\rfloor \in \{0,\dots,K-1\},
\label{eq:quantize}
\end{equation}
and dequantizes it when generating a structure at the bin center, refined by a learned sub-bin offset. The modulo in \cref{eq:quantize} places each coordinate on a circle of $K$ bins, so that a site near $f=1$ sits one step from a site near $f=0$.

\subsection{Canonicalization and Symmetry Augmentation}\label{sec:maskgxt_symmetry}

The same crystal can be written in many ways, since its unit-cell origin and atom ordering are arbitrary. MaskGXT trains the model to assign the same space group and Wyckoff tokens to all of these descriptions, using two online augmentations: a Euclidean-normalizer origin shift and an intra/inter-orbit permutation. Both are applied per sample during training, with the probabilities listed in \cref{tab:config}.

\paragraph{Euclidean-normalizer origin shift.}
The unit-cell origin is arbitrary, so each crystal admits many equivalent coordinate descriptions. Shifting the origin changes both the coordinates and the Wyckoff site of each atom, so we move them together. We draw a continuous origin shift $\mathbf{r}\sim\mathcal{U}[0,1)^3$, translate every atom's fractional coordinates by $\mathbf{r}$ modulo one, and relabel each Wyckoff letter to the value the shift induces, given by the normalizer coset whose representative is nearest to $\mathbf{r}$. Unlike \citet{kazeev2025wyckofftransformer}, who enumerate discrete coset representatives of the space group's Euclidean normalizer, our shift is continuous, exposing the model to every equivalent description. We precompute the normalizer translation cosets and their Wyckoff-letters for all 230 space groups, so relabeling is a table lookup at train time (\cref{alg:training}).

\paragraph{Intra/inter-orbit permutation.}
To exploit crystal symmetry, we order atoms by the canonical ordering of MCFlow~\citep{seong2026multimodal}, which groups sites into symmetry orbits and sorts them. This ordering still leaves equivalent descriptions, since the order among orbits sharing the same element and Wyckoff position, and of atoms within an orbit, is arbitrary. We train on these equivalent descriptions with its intra/inter-orbit permutation augmentation, which shuffles whole orbits and the atoms within each orbit. By permuting only within the orbit structure rather than over all $N!$ atom orders, it augments only symmetry-equivalent descriptions and outperforms full random permutation.

\subsection{Masked Learning Objective}
\label{sec:maskgxt_objective}

\paragraph{Ordinal, circular label smoothing.}
Treating the bins as unordered categories would discard the metric structure of coordinates and the adjacency of bin $0$ and bin $K{-}1$ on this circle. MaskGXT therefore replaces the one-hot target by an \emph{ordinal, circular} soft target~\citep{szegedy2016inception,muller2019labelsmoothing}: for a target bin $b$, the supervision over bin $b'$ is
\begin{equation}
p^{\star}_{b}(b') \;\propto\; \exp\!\left(-\frac{d(b,b')^2}{2\sigma^2}\right),
\qquad
d(b,b') = \min\bigl(|b-b'|,\; K-|b-b'|\bigr),
\label{eq:softtarget}
\end{equation}
where $\sigma$ controls the spread. The soft target gives most of its probability to the true bin and a little to nearby bins, so predicting a neighboring bin incurs less loss than predicting a far one. Lattice tokens use the same construction with the ordinary, non-circular distance, since lattice parameters do not wrap.

\paragraph{Masked cross-entropy.}
Following masked generative modeling~\citep{austin2021d3pm,chang2022maskgit}, training corrupts a crystal by independently replacing each token with a mask symbol with probability $t\sim\mathcal{U}(0,1)$, and trains a transformer $f_\theta$ to predict the original tokens at the masked positions. The loss is a masked cross-entropy against the per-token targets, summed over the four streams with weights $w_m$:
\begin{equation}
\mathcal{L}(\theta) \;=\; \mathbb{E}_{t,\,\mathcal{C},\,\mathcal{M}}
\sum_{m\in\{\mathrm{lat},\mathrm{coord},\mathrm{sg},\mathrm{wyk}\}}
w_m \, \frac{1}{|\mathcal{M}_m|}\sum_{i\in\mathcal{M}_m}
\operatorname{CE}\!\left(f_\theta(\mathcal{C}_{\setminus\mathcal{M}})_i,\; p^{\star}_{y_i}\right),
\label{eq:loss}
\end{equation}
where $\mathcal{C}$ is the tokenized crystal, $\mathcal{M}$ the set of masked positions, $\mathcal{C}_{\setminus\mathcal{M}}$ the sequence with those positions replaced by the mask symbol, $\mathcal{M}_m$ the masked positions within stream $m$, and $y_i$ the clean target at position $i$; the weights $w_m$ balance the four streams, and the space group and Wyckoff streams use one-hot targets.

\paragraph{Sub-bin coordinate refinement.}
While discretization provides a stable categorical representation, it limits geometric precision to half a bin width. To recover sub-bin accuracy, we attach a lightweight regression head that reads the trunk feature $\mathbf{h}$ at each lattice and coordinate position and predicts a bounded offset $\delta\in(-\tfrac12,\tfrac12)$. The offset is supervised by the true residual relative to the corresponding bin center using a smooth-$L_1$ loss, which is added to \cref{eq:loss}. Specifically,
\begin{equation}
\delta = \tfrac12\tanh\!\left(\mathbf{w}_{\mathrm{off}}^{\top}\mathbf{h}\right),
\quad
\delta^{\star} = K f - \bigl(q(f)+\tfrac12\bigr),
\quad
\mathcal{L}_{\mathrm{off}} = \mathbb{E}\,\rho\!\left(\delta-\delta^{\star}\right),
\label{eq:offset}
\end{equation}
where $f$ denotes the ground-truth fractional or lattice value, $\delta^{\star}$ is its residual from the center of bin $q(f)$, and $\rho$ is the smooth-$L_1$ penalty, averaged over unmasked lattice and coordinate positions. During decoding, MaskGXT reconstructs each continuous value by shifting the predicted bin center by this offset,
$\hat f = (q(f)+\tfrac12+\delta)/K$ (\cref{tab:arch_ablation}). The $\tfrac12\tanh$ bounds the offset to half a bin width, so the refined value is in its bin.

\begin{figure}[t]
  \centering
  \includegraphics[width=\textwidth]{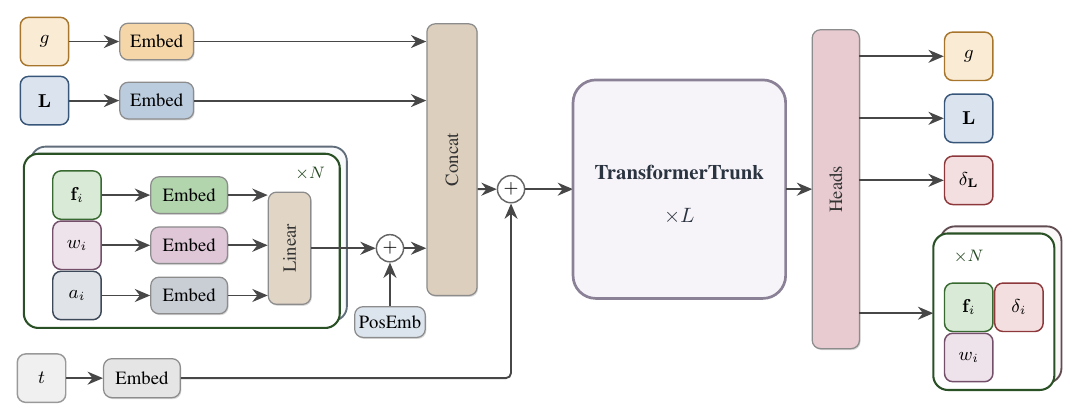}
  \caption{\textbf{MaskGXT architecture.}
  A per-site linear projection fuses the five token embeddings of each site
  (three coordinate, one Wyckoff, one atom-type) into one atom-wise feature; with
  the space group, lattice, and mask-ratio features, the sequence passes through a
  bidirectional $\operatorname{TransformerTrunk}$, and linear heads read off the
  per-stream predictions. A lightweight regression head additionally predicts a
  bounded sub-bin offset ($\delta_{\mathbf{L}}$ for the lattice, $\delta_i$ per
  coordinate site) that refines the discrete prediction (\cref{eq:offset}).}
  \label{fig:arch}
\end{figure}
\subsection{Model Architecture}
\label{sec:maskgxt_architecture}

The model architecture consists of three parts: (1) a tokenization scheme, (2) a Transformer trunk, and (3) prediction heads (\cref{fig:arch}). After embedding the $1{+}6{+}5N$ tokens, a per-site linear projection $\mathbf{W}_{\mathrm{site}}$ fuses the five token embeddings of each site into one atom-wise feature, yielding a length-$1{+}6{+}N$ sequence of one space group, six lattice, and $N$ atom-wise features. Each atom-wise feature further adds a learned positional encoding over the canonical site order and a mask-ratio embedding before the bidirectional Transformer. The sequence is processed by $L$ layers of $\operatorname{TransformerTrunk}$, and linear heads then read from the shared trunk to produce categorical logits for the space group, lattice, coordinate, and Wyckoff streams. Specifically, the forward pass is given by:
\begin{align}
&\mathbf{h}_g = \mathbf{e}_g(g),
\qquad
\mathbf{h}^{\mathrm{lat}}_j = \mathbf{e}(\ell_j) + \mathbf{a}^{\mathrm{lat}}_j,
\notag\\
&\mathbf{h}^{\mathrm{site}}_i = \mathbf{W}_{\mathrm{site}}\!\left[\,\mathbf{e}(q(x_i)){+}\mathbf{a}_x;\ \mathbf{e}(q(y_i)){+}\mathbf{a}_y;\ \mathbf{e}(q(z_i)){+}\mathbf{a}_z;\ \mathbf{e}_Z(a_i);\ \mathbf{e}_w(w_i)\,\right] + \mathbf{p}_i,
\notag\\
&\mathbf{H}^{(0)} = \left[\,\mathbf{h}_g;\ \mathbf{h}^{\mathrm{lat}}_{1:6};\ \mathbf{h}^{\mathrm{site}}_{1:N}\,\right] + \mathbf{1}\,\boldsymbol{\tau}_t^{\top},
\qquad
\boldsymbol{\tau}_t = \mathbf{W}_t\,\operatorname{SiLU}\!\left(\operatorname{SinEmb}(t)\right),
\notag\\
&\mathbf{H}^{(L)} = \operatorname{TransformerTrunk}\!\left(\mathbf{H}^{(0)}\right),
\qquad
f_\theta(\mathcal{C}) = \bigl(\operatorname{Head}_m(\mathbf{H}^{(L)})\bigr)_{m\in\{\mathrm{sg},\,\mathrm{lat},\,\mathrm{lat\text{-}off},\,\mathrm{coord},\,\mathrm{coord\text{-}off},\,\mathrm{wyk}\}},
\label{eq:forward}
\end{align}
where $\mathcal{C}$ is the token sequence, $\ell_j$ the $j$-th lattice bin token, $\mathbf{e}$ the shared bin embedding, $\mathbf{a}^{\mathrm{lat}}_j$ and $\mathbf{a}_{\{x,y,z\}}$ per-axis encodings, and $\mathbf{e}_g,\mathbf{e}_Z,\mathbf{e}_w$ the space group, atom-type, and Wyckoff embeddings. The positional encoding factors as $\mathbf{p}_i=\mathbf{p}^{\mathrm{elem}}_{k_i}+\mathbf{p}^{\mathrm{within}}_{r_i}$, a learned element-block encoding at the rank $k_i$ of atom $i$'s element in the canonical order plus a within-element encoding at its rank $r_i$ within that element. Each $\operatorname{Head}_m$ produces categorical logits for the four token streams and bounded $\tfrac12\tanh$ sub-bin offsets for the lattice and coordinate positions (\cref{sec:maskgxt_objective}).

\paragraph{Transformer trunk.}
The $\operatorname{TransformerTrunk}$ of \cref{eq:forward} is a Pre-LN Transformer~\citep{xiong2020layer} with QK-normalized self-attention~\citep{dehghani2023scaling}. It consists of $L$ blocks, each applying multi-head self-attention followed by a SwiGLU feedforward network~\citep{shazeer2020glu}. Specifically, for a block input sequence $\mathbf{x}$, let $\tilde{\mathbf{x}}=\operatorname{LN}(\mathbf{x})$ and $\mathbf{q},\mathbf{k},\mathbf{v}=\operatorname{proj}(\tilde{\mathbf{x}})$ denote the per-head query, key, and value projections, and $\mathbf{W}_O$ the attention output projection. Each block updates
\begin{align}
\mathbf{x} &\leftarrow \mathbf{x} + \mathbf{W}_O\!\left(\softmax\!\left(s_h\,\hat{\mathbf{q}}\hat{\mathbf{k}}^{\top}\right)\mathbf{v}\right),
\qquad
\hat{\mathbf{q}}=\frac{\mathbf{q}}{\lVert\mathbf{q}\rVert},\ \ \hat{\mathbf{k}}=\frac{\mathbf{k}}{\lVert\mathbf{k}\rVert},
\\
\mathbf{x} &\leftarrow \mathbf{x} + \mathbf{W}_2\!\left(\operatorname{SiLU}\!\left(\mathbf{W}_1\,\operatorname{LN}(\mathbf{x})\right)\odot \mathbf{W}_3\,\operatorname{LN}(\mathbf{x})\right).
\end{align}
QK-normalization normalizes queries and keys along the head dimension, bounding their dot products, while a learnable per-head temperature $s_h$ replaces the standard $1/\sqrt{d_h}$ scaling. The SwiGLU hidden width is $\tfrac{2}{3}\cdot 4d$, rounded to a multiple of $64$.

\subsection{Sampling}
\label{sec:maskgxt_decoding}

\paragraph{Space group stratified sampling.}
A composition may form several polymorphs, so MaskGXT samples a diverse set of candidates by branching over space groups. For a requested budget of $S$ candidates per composition, it computes the space group posterior from the fully masked sequence $\mathcal{C}_{\mathrm{mask}}$ with only the composition observed,
$
p_\theta(g \mid A)=\softmax\bigl(f_\theta(\mathcal{C}_{\mathrm{mask}})_{\mathrm{sg}}\bigr)$,
selects the $S$ highest-posterior space groups, and creates one sampling branch for each selected $g$. In each branch, the space group token is fixed to $g$ and the remaining lattice, coordinate, and Wyckoff tokens are decoded by the greedy procedure below (\cref{alg:polymorph}). When every remaining space group has posterior mass below a small threshold, the branch instead samples its space group and tokens i.i.d.\ (\cref{app:config}).

\paragraph{Confidence-ranked greedy decoding.}
Within each branch, MaskGXT returns the most probable structure by replacing stochastic token sampling with deterministic token selection. Starting from the fully masked sequence with $\mathbf{A}$ and the branch space group given, it iteratively predicts posterior distributions over the masked token streams, fills the highest-confidence positions using token-wise \emph{argmax}, and re-masks the remaining positions according to the decoding schedule (\cref{alg:sampling}). Because the output space is discretized, this greedy rule is well defined over a finite vocabulary for lattice, coordinate, space group, and Wyckoff tokens. It therefore returns a high-confidence deterministic structure, in contrast to diffusion- and flow-based samplers whose outputs are obtained by integrating or sampling a continuous generative process. This improves single-structure accuracy in our ablation (\cref{tab:decoding_ablation}).

\section{Experiments}
\label{sec:experiments}

The experiments answer two questions: (1) whether the discrete masked generative formulation outperforms continuous diffusion- and flow-based CSP models under the standard one-to-one evaluation protocol, and (2) whether the components of MaskGXT improve geometric accuracy and polymorph coverage.

\subsection{Datasets}
\label{sec:exp_datasets}

We evaluate on two standard inorganic-crystal CSP benchmarks: \emph{MP-20}~\citep{xie2021crystal} and \emph{MPTS-52}~\citep{baird2024mpts}. MP-20 contains materials from the Materials Project~\citep{jain2013commentary} with at most 20 atoms per unit cell and formation energies within 0.08~eV/atom of the convex hull. MPTS-52 extends the setting to structures with up to 52 atoms per cell and uses a chronological split by publication year. We follow the standard 60/20/20 train/validation/test partition for both datasets.

We additionally evaluate on the MP-20 polymorph split of \citet{martirossyan2025metre}. Unlike the standard random structure split, this split assigns all polymorphs of a composition to the same subset. Its test set therefore retains compositions with multiple reference polymorphs, enabling a direct evaluation of polymorph coverage under METRe.

\subsection{Evaluation Metrics}
\label{sec:exp_metrics}

We compare generated and reference structures using \texttt{StructureMatcher}~\citep{ong2013python}, following the CSP evaluation protocol of \citet{jiao2023crystal,miller2024flowmm}. Unless otherwise stated, matching uses tolerances $\mathrm{ltol}=0.3$, $\mathrm{stol}=0.5$, and $\mathrm{angle\_tol}=10^\circ$.

For standard one-to-one CSP evaluation, each test reference is paired with one generated structure of the same composition. We report match rate (MR), defined as the fraction of test references that are matched, and RMSE over matched pairs. We compute both metrics in two forms: an \emph{unfiltered} score, applied directly to the generated set, and a \emph{validity-filtered} score, which treats a generated structure as unmatched if it fails standard chemical or structural validity checks.

For polymorph-aware evaluation, we report METRe~\citep{martirossyan2025metre}. METRe uses the same matching primitive but pools generated structures by composition: a reference structure is counted as matched if any generated structure of the same composition matches it. We also report the corrected RMSE (cRMSE), which assigns each matched reference its best matched RMSE and each unmatched reference the tolerance cap $\mathrm{stol}$, thereby summarizing both coverage and geometric precision. All final metrics are evaluated on held-out test splits disjoint from the validation split used by the agent during search (\cref{sec:agent}).

\subsection{Baselines and MaskGXT Protocol}
\label{sec:exp_protocol}

We compare MaskGXT against DiffCSP~\citep{jiao2023crystal}, FlowMM~\citep{miller2024flowmm}, OMatG~\citep{hoellmer2025omat}, Crystalite~\citep{veljkovic2026crystalite}, and MCFlow~\citep{seong2026multimodal}. For OMatG, we use the linear-ODE variant; for MCFlow, we use the base variant. METRe values are reproduced from \citet{martirossyan2025metre} and, for MCFlow, from \citet{seong2026multimodal}.

MaskGXT is evaluated as a composition-conditioned CSP model: given only the test composition, it generates crystal structures by confidence-ranked masked parallel decoding over the token sequence described in \cref{sec:maskgxt}. Unless otherwise specified, we follow the one-to-one protocol used by the baselines and generate one structure per test reference for MR and RMSE evaluation. For METRe, we use the space group stratified sampling protocol in \cref{sec:maskgxt}, which spreads generations over high-posterior space groups to improve coverage of multiple polymorphs of the same composition. The reported MaskGXT model is an approximately 248M-parameter transformer; its architecture and hyperparameters are given in \cref{tab:config}. The final model trains for up to $3000$ epochs with early stopping on validation METRe.

\subsection{Standard CSP Performance}
\label{sec:exp_standard_csp}

\begin{table}[t]
\centering
\caption{
\textbf{Crystal structure prediction on the MP-20 and MPTS-52 test sets.}
Each cell gives the match rate and matched-structure RMSE as unfiltered/filtered values.
MaskGXT uses greedy decoding for this evaluation.
MaskGXT attains the best match rate on both MP-20 and MPTS-52.
\textbf{Bold} marks the best value in each column.
$^{\dagger}$Reproduced by running the official Crystalite code with the configuration reported in~\citet{veljkovic2026crystalite}; the unmarked Crystalite row is taken directly from that paper.
}
\label{tab:csp}
\small
\setlength{\tabcolsep}{5pt}
\begin{tabular}{lcccc}
\toprule
\multirow{2}{*}{Model}
& \multicolumn{2}{c}{MP-20} & \multicolumn{2}{c}{MPTS-52} \\
\cmidrule(lr){2-3} \cmidrule(lr){4-5}
& MR (\%) $\uparrow$ & RMSE $\downarrow$ & MR (\%) $\uparrow$ & RMSE $\downarrow$ \\
\midrule
DiffCSP                          & 57.82 / 52.51 & 0.0627 / 0.0600 & 15.79 / 14.29 & 0.1533 / 0.1489 \\
FlowMM                           & 66.22 / 59.98 & 0.0661 / 0.0629 & 22.29 / 20.28 & 0.1541 / 0.1486 \\
OMatG                            & 69.83 / 63.75 & 0.0741 / 0.0720 & 27.38 / 25.15 & 0.1970 / 0.1931 \\
MCFlow                           & 69.23 / 63.14 & 0.0663 / 0.0611 & 28.77 / 26.46 & 0.1610 / 0.1577 \\
Crystalite                       & \phantom{00.00} / 66.05 & \phantom{0.0000} / 0.0329 & \phantom{00.00} / 31.49 & \phantom{0.0000} / \textbf{0.0701} \\
Crystalite$^{\dagger}$           & 69.64 / 63.46 & 0.0412 / 0.0397 & 27.26 / 24.98 & 0.1181 / 0.1180 \\
\midrule
\textbf{MaskGXT}                 & \textbf{73.79 / 67.06} & \textbf{0.0330 / 0.0325} & \textbf{36.75 / 33.34} & \textbf{0.1004} / 0.0975 \\
\bottomrule
\end{tabular}
\end{table}

\Cref{tab:csp} reports standard one-to-one CSP performance on MP-20 and MPTS-52. MaskGXT achieves the highest match rate on both datasets, under both the unfiltered and validity-filtered definitions, and also attains the lowest matched-pair RMSE among the compared methods. On MP-20 this RMSE is essentially on par with our reproduced Crystalite, while the improvement is particularly large on the harder MPTS-52 split, in both match rate and RMSE. This suggests that the discrete masked generative formulation generalizes beyond the MP-20 development setting, and that discretizing coordinates, when combined with sub-bin offset regression, does not sacrifice geometric precision.

\subsection{Polymorph-aware Evaluation}
\label{sec:exp_polymorph}

\begin{table}[t]
\centering
\caption{
\textbf{Crystal structure prediction on the METRe metric.}
All values are on the held-out test set. Following the METRe protocol, each
composition is given a sampling budget equal to its number of reference
structures in the test set. MaskGXT uses space group stratified sampling and
leads on both splits. \textbf{Bold} marks the best value in each column.
$^{\dagger}$Reproduced by running the official Crystalite code with the configuration reported in~\citet{veljkovic2026crystalite}, as METRe is not reported in that paper.
}
\label{tab:metre}
\small
\setlength{\tabcolsep}{5pt}
\begin{tabular}{lcccc}
\toprule
\multirow{2}{*}{Model} & \multicolumn{2}{c}{MP-20} & \multicolumn{2}{c}{MP-20 polymorph split} \\
\cmidrule(lr){2-3} \cmidrule(lr){4-5}
& METRe (\%) $\uparrow$ & cRMSE $\downarrow$ & METRe (\%) $\uparrow$ & cRMSE $\downarrow$ \\
\midrule
DiffCSP                  & 58.80 & 0.244 & 53.14 & 0.279 \\
FlowMM                   & 67.00 & 0.210 & 65.18 & 0.226 \\
OMatG                    & 66.00 & 0.208 & 70.50 & 0.187 \\
MCFlow                   & 69.70 & 0.200 & 70.70 & 0.195 \\
Crystalite$^{\dagger}$   & 70.45 & 0.178 & 70.87 & 0.174 \\
\midrule
\textbf{MaskGXT}         & \textbf{74.78} & \textbf{0.152} & \textbf{79.06} & \textbf{0.132} \\
\bottomrule
\end{tabular}
\end{table}

\Cref{tab:metre} evaluates polymorph coverage with METRe. MaskGXT obtains the best METRe and cRMSE on both the standard MP-20 split and the MP-20 polymorph split, with the largest advantage on the polymorph split. This pattern matches the design of the sampler: because MaskGXT predicts the space group as an explicit token, space group stratified sampling allocates different generations to distinct high-posterior space groups, improving coverage of compositions with multiple polymorphs. The standard MP-20 split has little polymorph multiplicity and leaves less coverage to recover, so the gain there is smaller; the larger gain on the polymorph split reflects genuine recovery of multiple structures per composition.

\begin{table}[t]
\centering
\caption{
\textbf{Crystal structure prediction under uniform sampling.}
At test time the number of polymorphs per composition is unknown, so each model
draws a fixed $S=2$ structures for every unique composition. METRe is measured
against the full reference set. \textbf{Bold} marks the best value in each column.
$^{\dagger}$Reproduced by running the official Crystalite code as in~\citet{veljkovic2026crystalite}.
}
\label{tab:uniform}
\small
\setlength{\tabcolsep}{5pt}
\begin{tabular}{lcccc}
\toprule
\multirow{2}{*}{Model} & \multicolumn{2}{c}{MP-20} & \multicolumn{2}{c}{MP-20 polymorph split} \\
\cmidrule(lr){2-3} \cmidrule(lr){4-5}
& METRe (\%) $\uparrow$ & cRMSE $\downarrow$ & METRe (\%) $\uparrow$ & cRMSE $\downarrow$ \\
\midrule
OMatG                    & 72.55 & 0.187 & 68.23 & \pscrmseomatg \\
Crystalite$^{\dagger}$   & 72.83 & 0.164 & 68.44 & \pscrmsecryst \\
\midrule
\textbf{MaskGXT}         & \textbf{76.53} & \textbf{0.143} & \textbf{75.00} & \textbf{\pscrmsemask} \\
\bottomrule
\end{tabular}
\end{table}

\textbf{Uniform sampling.} The METRe protocol in \Cref{tab:metre} gives each composition a budget equal to its number of reference structures, which is unavailable at deployment time. We therefore evaluate a realistic setting where the number of polymorphs is unknown and every composition receives the same budget of $S$ structures. At $S=2$ (\Cref{tab:uniform}), MaskGXT remains the strongest method on both splits, and its margin over the baselines widens on the polymorph split, where stratified sampling spends the uniform budget on distinct space groups rather than redundant draws. This lead holds as $S$ varies from one to five (\cref{app:sweep}).

\subsection{Ablation Studies}
\label{sec:exp_ablations}

\begin{table}[t]
\centering
\caption{\textbf{Architecture ablation.} Each row removes one training-time component from MaskGXT. MR (unfiltered/filtered) and RMSE use greedy decoding; METRe and cRMSE additionally use space group stratified sampling. Without symmetry tokens, space group stratification is unavailable, so that row's METRe uses unstratified greedy decoding. \textbf{Bold} marks the best value in each column.}
\label{tab:arch_ablation}
\small
\setlength{\tabcolsep}{3pt}
\begin{tabular}{lcccccc}
\toprule
\multirow{2}{*}{Model} & \multicolumn{4}{c}{MP-20} & \multicolumn{2}{c}{MP-20 polymorph split} \\
\cmidrule(lr){2-5} \cmidrule(lr){6-7}
& MR & RMSE & METRe & cRMSE & METRe & cRMSE \\
\midrule
\textbf{MaskGXT}                   & 73.79 / 67.06 & \textbf{0.0330 / 0.0325} & \textbf{74.78} & \textbf{0.1518} & \textbf{79.06} & \textbf{0.1322} \\
\quad $-$ label smoothing          & 72.16 / 65.45 & 0.0396 / 0.0388 & 73.16 & 0.1639 & 77.23 & 0.1457 \\
\quad $-$ symmetry tokens          & 72.08 / 65.30 & 0.0408 / 0.0392 & 72.51 & 0.1674 & 74.98 & 0.1551 \\
\quad $-$ offset regression        & \textbf{73.93 / 67.11} & 0.0518 / 0.0511 & 74.65 & 0.1607 & 78.70 & 0.1410 \\
\bottomrule
\end{tabular}
\end{table}
\begin{table}[t]
\centering
\caption{\textbf{Decoder ablation.} Greedy decoding and space group stratification trade match rate (MR) against METRe: greedy maximizes the single-structure match rate, while stratification maximizes polymorph coverage. Each MR and RMSE cell is unfiltered\,/\,validity-filtered. \textbf{Bold} marks the best value in each column.}
\label{tab:decoding_ablation}
\small
\setlength{\tabcolsep}{3pt}
\begin{tabular}{lcccccc}
\toprule
\multirow{2}{*}{Model} & \multicolumn{4}{c}{MP-20} & \multicolumn{2}{c}{MP-20 polymorph split} \\
\cmidrule(lr){2-5} \cmidrule(lr){6-7}
& MR & RMSE & METRe & cRMSE & METRe & cRMSE \\
\midrule
\textbf{MaskGXT}                 & 73.76 / \textbf{67.06} & 0.0333 / 0.0326 & \textbf{74.78} & \textbf{0.1518} & \textbf{79.06} & \textbf{0.1322} \\
\quad $-$ space group stratification & \textbf{73.79 / 67.06} & \textbf{0.0330 / 0.0325} & 73.84 & 0.1552 & 75.50 & 0.1473 \\
\quad $-$ greedy decoding         & 73.12 / 66.33 & 0.0517 / 0.0498 & 74.19 & 0.1678 & 78.39 & 0.1508 \\
\quad $-$ both                   & 73.28 / 66.36 & 0.0541 / 0.0517 & 74.17 & 0.1698 & 76.88 & 0.1586 \\
\bottomrule
\end{tabular}
\end{table}

We perform ablations to isolate the effect of MaskGXT's training-time and decoding-time components. \Cref{tab:arch_ablation} removes one training-time component at a time. Label smoothing and symmetry tokens both improve match rate and METRe, showing that periodic ordinal supervision and explicit crystallographic tokens contribute to structural accuracy and coverage. Removing symmetry tokens is especially harmful on the polymorph split, because space group-stratified sampling is no longer available. Removing offset regression has little effect on match rate but substantially increases RMSE and cRMSE, confirming that the offset head primarily improves geometric precision.

\Cref{tab:decoding_ablation} ablates the decoding strategy. Removing space group stratification causes the largest METRe drop on the polymorph split, confirming that spreading generations across high-posterior space groups is the main mechanism behind MaskGXT's improved polymorph coverage. Removing greedy decoding lowers METRe and worsens cRMSE, indicating that confidence-ranked argmax decoding produces more accurate high-posterior structures than stochastic decoding. Removing both components recovers a plain stochastic decoder without stratification and yields the lowest METRe, so the two decoder components contribute complementary gains.
\section{Conclusion}
\label{sec:conclusion}

We present MaskGXT, a masked generative crystal transformer that combines periodicity-aware coordinate discretization, ordinal label smoothing, tokenized crystallographic symmetry, and confidence-ranked greedy decoding. Importantly, MaskGXT is developed by an AI co-scientist loop that transfers masked generative modeling from computer vision to materials science and refines it with sparse, targeted human steering. This process produces key domain adaptations, including symmetry tokens, space group stratified sampling, and sub-bin coordinate refinement, which together improve structural accuracy and polymorph coverage. MaskGXT attains state-of-the-art match rate on both MP-20 and MPTS-52, and substantially improves METRe on the MP-20 polymorph split.

\paragraph{Scope and limitations.} The success of the AI co-scientist loop in CSP can be understood through the structure of the discovery problem. CSP provides a rare combination of fixed datasets, executable model code, fast training and evaluation cycles, and a validation metric that is closely aligned with the final task. These properties turn algorithm design into a high-throughput empirical feedback problem: the agent can propose a modeling principle, instantiate it in code, train it under a controlled budget, and receive a reliable signal on whether the change improves the algorithm. This signal is budget-dependent: our fixed compute budget penalizes methods that need more compute to converge. More broadly, the loop is harder to close in domains with physical experiments, long training cycles, or weak proxy objectives, which may instead require surrogate models, stronger human steering, or more structured evaluation.

\paragraph{Future directions.}
The same agent and tokenization extend naturally to related crystal generation tasks, including de novo generation and structure-conditioned atom-type generation within a single model. More broadly, this work highlights several requirements for making AI co-scientists more generally useful. 

First, future systems must address the human attention bottleneck in agentic research. A single search run can produce many hypotheses, code variants, logs, plots, failed attempts, and partial analyses, most of which are too raw for a human scientist to inspect. Rather than exposing this stream directly, an AI co-scientist should maintain a compact research state: which ideas have been tried, why they succeeded or failed, which mechanisms remain unexplored, and which decisions require human judgment. This is also a goal-specification problem, because the human should be able to revise the scientific objective at a conceptual level while the agent translates that revision into coordinated parallel research threads with preserved provenance. 

Second, agent search needs more principled proxy evaluation. Rather than selecting candidates only by short-budget runs, future systems should develop validation signals whose rankings extrapolate to larger budgets, for example through scaling-law analyses over model size, training time, data size, and sampling budget. 

Third, the search loop should expand beyond metric-optimizing experiments. Scientific algorithm discovery also requires exploratory and analytic experiments: mapping failure modes, inspecting generated samples, testing mechanistic hypotheses, designing ablations, and deciding which empirical signals are trustworthy. Equipping agents with these capabilities would move them from automated implementation search toward a fuller research workflow for transferring ideas across chemistry, biology, materials science, and other domains.

\clearpage
\bibliography{references}

@inproceedings{chang2022maskgit,
  title={{MaskGIT}: Masked Generative Image Transformer},
  author={Chang, Huiwen and Zhang, Han and Jiang, Lu and Liu, Ce and Freeman, William T.},
  booktitle={IEEE/CVF Conference on Computer Vision and Pattern Recognition (CVPR)},
  pages={11315--11325},
  year={2022}
}

@article{butler2018machine,
  title={Machine learning for molecular and materials science},
  author={Butler, Keith T and Davies, Daniel W and Cartwright, Hugh and Isayev, Olexandr and Walsh, Aron},
  journal={Nature},
  volume={559},
  number={7715},
  pages={547--555},
  year={2018},
  publisher={Nature Publishing Group}
}

@article{jain2013commentary,
  title={Commentary: The Materials Project: A materials genome approach to accelerating materials innovation},
  author={Jain, Anubhav and Ong, Shyue Ping and Hautier, Geoffroy and Chen, Wei and Richards, William Davidson and Dacek, Stephen and Cholia, Shreyas and Gunter, Dan and Skinner, David and Ceder, Gerbrand and others},
  journal={APL materials},
  volume={1},
  number={1},
  pages={011002},
  year={2013},
  publisher={American Institute of PhysicsAIP}
}

@article{ong2013python,
  title={Python Materials Genomics (pymatgen): A robust, open-source python library for materials analysis},
  author={Ong, Shyue Ping and Richards, William Davidson and Jain, Anubhav and Hautier, Geoffroy and Kocher, Michael and Cholia, Shreyas and Gunter, Dan and Chevrier, Vincent L and Persson, Kristin A and Ceder, Gerbrand},
  journal={Computational Materials Science},
  volume={68},
  pages={314--319},
  year={2013},
  publisher={Elsevier}
}

@article{oganov2019structure,
  title={Structure prediction drives materials discovery},
  author={Oganov, Artem R and Pickard, Chris J and Zhu, Qiang and Needs, Richard J},
  journal={Nature Reviews Materials},
  volume={4},
  number={5},
  pages={331--348},
  year={2019},
  publisher={Nature Publishing Group}
}

@inproceedings{xie2021crystal,
  title={Crystal diffusion variational autoencoder for periodic material generation},
  author={Xie, Tian and Fu, Xiang and Ganea, Octavian-Eugen and Barzilay, Regina and Jaakkola, Tommi},
  booktitle={International Conference on Learning Representations},
  year={2022}
}

@inproceedings{jiao2023crystal,
	title        = {Crystal Structure Prediction by Joint Equivariant Diffusion},
	author       = {Jiao, Rui and Huang, Wenbing and Lin, Peijia and Han, Jiaqi and Chen, Pin and Lu, Yutong and Liu, Yang},
	year         = 2023,
	booktitle    = {Advances in Neural Information Processing Systems},
	url          = {https://arxiv.org/abs/2309.04475},
	eprint       = {2309.04475},
	archiveprefix = {arXiv},
	primaryclass = {cond-mat.mtrl-sci}
}

@inproceedings{jiao2024space,
	title        = {Space Group Constrained Crystal Generation},
	author       = {Rui Jiao and Wenbing Huang and Yu Liu and Deli Zhao and Yang Liu},
	year         = 2024,
	booktitle    = {International Conference on Learning Representations},
	url          = {https://arxiv.org/abs/2402.03992},
	eprint       = {2402.03992},
	archiveprefix = {arXiv},
	primaryclass = {cs.LG}
}

@article{zeni2023mattergen,
	title        = {{MatterGen}: a generative model for inorganic materials design},
	author       = {Zeni, Claudio and Pinsler, Robert and Z{\"u}gner, Daniel and Fowler, Andrew and Horton, Matthew and Fu, Xiang and Shysheya, Sasha and Crabb{\'e}, Jonathan and Sun, Lixin and Smith, Jake and others},
	year         = 2025,
	journal      = {Nature},
	volume       = 639,
	pages        = {624--632},
	url          = {https://arxiv.org/abs/2312.03687},
	eprint       = {2312.03687},
	archiveprefix = {arXiv},
	primaryclass = {cond-mat.mtrl-sci}
}

@misc{baird2024mpts,
	title        = {sparks-baird/matbench-genmetrics},
	author       = {Sterling G. Baird and Hasan Muhammad Sayeed and Joseph Montoya and Taylor D. Sparks},
	year         = 2024,
	note         = {[Accessed 03-05-2024]},
	howpublished = {\url{https://github.com/sparks-baird/matbench-genmetrics}}
}

@inproceedings{miller2024flowmm,
	title        = {{FlowMM}: Generating Materials with {R}iemannian Flow Matching},
	author       = {Benjamin Kurt Miller and Ricky T. Q. Chen and Anuroop Sriram and Brandon M Wood},
	year         = 2024,
	booktitle    = {International Conference on Machine Learning},
	url          = {https://arxiv.org/abs/2406.04713},
	eprint       = {2406.04713},
	archiveprefix = {arXiv},
	primaryclass = {cs.LG}
}

@inproceedings{levy2025symmcd,
	title        = {SymmCD: Symmetry-Preserving Crystal Generation with Diffusion Models},
	author       = {Daniel Levy and Siba Smarak Panigrahi and Sékou-Oumar Kaba and Qiang Zhu and Kin Long Kelvin Lee and Mikhail Galkin and Santiago Miret and Siamak Ravanbakhsh},
	year         = 2025,
	booktitle    = {International Conference on Learning Representations},
	url          = {https://arxiv.org/abs/2502.03638},
	eprint       = {2502.03638},
	archiveprefix = {arXiv},
	primaryclass = {cond-mat.mtrl-sci}
}

@article{zhu2024wycryst,
  title={{WyCryst}: {W}yckoff Inorganic Crystal Generator Framework},
  author={Zhu, Ruiming and Nong, Wei and Yamazaki, Shuya and Hippalgaonkar, Kedar},
  journal={Matter},
  volume={7},
  number={10},
  pages={3469--3488},
  year={2024},
  publisher={Elsevier}
}

@inproceedings{kazeev2025wyckofftransformer,
	title        = {Wyckoff Transformer: Generation of Symmetric Crystals},
	author       = {Nikita Kazeev and Wei Nong and Ignat Romanov and Ruiming Zhu and Andrey Ustyuzhanin and Shuya Yamazaki and Kedar Hippalgaonkar},
	year         = 2025,
	booktitle    = {International Conference on Machine Learning},
	url          = {https://arxiv.org/abs/2503.02407},
	eprint       = {2503.02407},
	archiveprefix = {arXiv},
	primaryclass = {cond-mat.mtrl-sci}
}

@inproceedings{hoellmer2025omat,
	title        = {Open Materials Generation with Stochastic Interpolants},
	author       = {Philipp Hoellmer and Thomas Egg and Maya M. Martirossyan and Eric Fuemmeler and Zeren Shui and Amit Gupta and Pawan Prakash and Adrian Roitberg and Mingjie Liu and George Karypis and Mark Transtrum and Richard G. Hennig and Ellad B. Tadmor and Stefano Martiniani},
	year         = 2025,
	booktitle    = {International Conference on Machine Learning},
	url          = {https://arxiv.org/abs/2502.02582},
	eprint       = {2502.02582},
	archiveprefix = {arXiv},
	primaryclass = {cs.LG}
}

@article{luo2025crystalflow,
  title={{CrystalFlow}: A Flow-based Generative Model for Crystalline Materials},
  author={Luo, Xiaoshan and Wang, Zhenyu and Wang, Qingchang and Shao, Xuechen and Lv, Jian and Wang, Lei and Wang, Yanchao and Ma, Yanming},
  journal={Nature Communications},
  volume={16},
  number={1},
  pages={9267},
  year={2025},
  publisher={Nature Publishing Group UK London}
}

@inproceedings{kelvinius2025wyckoffdiff,
  title={{WyckoffDiff}--A Generative Diffusion Model for Crystal Symmetry},
  author={Kelvinius, Filip Ekstr{\"o}m and Andersson, Oskar B and Parackal, Abhijith S and Qian, Dong and Armiento, Rickard and Lindsten, Fredrik},
  booktitle={Forty-second International Conference on Machine Learning},
  year={2025},
  eprint={2502.06485},
  archiveprefix={arXiv},
  primaryclass={cond-mat.mtrl-sci}
}

@inproceedings{martirossyan2025metre,
  title={All that structure matches does not glitter},
  author={Martirossyan, Maya M and Egg, Thomas and Hoellmer, Philipp and Karypis, George and Transtrum, Mark and Roitberg, Adrian and Liu, Mingjie and Hennig, Richard G and Tadmor, Ellad B and Martiniani, Stefano},
  booktitle={Advances in Neural Information Processing Systems},
  year={2025}
}

@article{crystalformer2025,
  title={Space Group Informed Transformer for Crystalline Materials Generation},
  author={Cao, Zhendong and Luo, Xiaoshan and Lv, Jian and Wang, Lei},
  journal={Science Bulletin},
  year={2025}
}

@article{gottweis2025coscientist,
  title={Towards an {AI} Co-Scientist},
  author={Gottweis, Juraj and Weng, Wei-Hung and Daryin, Alexander and Tu, Tao and Palepu, Anil and Sirkovic, Petar and Myaskovsky, Artiom and Weissenberger, Felix and Rong, Keran and Tanno, Ryutaro and Saab, Khaled and Popovici, Dan and Blum, Jacob and Zhang, Fan and Chou, Katherine and Hassidim, Avinatan and Gokturk, Burak and Vahdat, Amin and Kohli, Pushmeet and Matias, Yossi and Carroll, Andrew and Kulkarni, Kavita and Tomasev, Nenad and Guan, Yuan and Natarajan, Vivek},
  journal={arXiv preprint arXiv:2502.18864},
  year={2025}
}

@article{lu2024aiscientist,
  title={The {AI} Scientist: Towards Fully Automated Open-Ended Scientific Discovery},
  author={Lu, Chris and Lu, Cong and Lange, Robert Tjarko and Foerster, Jakob and Clune, Jeff and Ha, David},
  journal={arXiv preprint arXiv:2408.06292},
  year={2024}
}

@article{wei2025agentic,
  title={From {AI} for Science to Agentic Science: A Survey on Autonomous Scientific Discovery},
  author={Wei, Jiaqi and Yang, Yuejin and Zhang, Xiang and others},
  journal={arXiv preprint arXiv:2508.14111},
  year={2025}
}

@article{yamada2025aiscientistv2,
  title={The {AI} Scientist-v2: Workshop-Level Automated Scientific Discovery via Agentic Tree Search},
  author={Yamada, Yutaro and Lange, Robert Tjarko and Lu, Cong and Hu, Shengran and Lu, Chris and Foerster, Jakob and Clune, Jeff and Ha, David},
  journal={arXiv preprint arXiv:2504.08066},
  year={2025}
}

@article{novikov2025alphaevolve,
  title={{AlphaEvolve}: A Coding Agent for Scientific and Algorithmic Discovery},
  author={Novikov, Alexander and V{\~u}, Ng{\^a}n and Eisenberger, Marvin and Dupont, Emilien and Huang, Po-Sen and Wagner, Adam Zsolt and Shirobokov, Sergey and Kozlovskii, Borislav and Ruiz, Francisco J. R. and Mehrabian, Abbas and Kumar, M. Pawan and See, Abigail and Chaudhuri, Swarat and Holland, George and Davies, Alex and Nowozin, Sebastian and Kohli, Pushmeet and Balog, Matej},
  journal={arXiv preprint arXiv:2506.13131},
  year={2025}
}

@article{romeraparedes2024funsearch,
  title={Mathematical Discoveries from Program Search with Large Language Models},
  author={Romera-Paredes, Bernardino and Barekatain, Mohammadamin and Novikov, Alexander and Balog, Matej and Kumar, M. Pawan and Dupont, Emilien and Ruiz, Francisco J. R. and Ellenberg, Jordan S. and Wang, Pengming and Fawzi, Omar and Kohli, Pushmeet and Fawzi, Alhussein},
  journal={Nature},
  volume={625},
  number={7995},
  pages={468--475},
  year={2024},
  publisher={Nature Publishing Group}
}

@article{jiang2025aide,
  title={{AIDE}: {AI}-Driven Exploration in the Space of Code},
  author={Jiang, Zhengyao and Schmidt, Dominik and Srikanth, Dhruv and Xu, Dixing and Kaplan, Ian and Jacenko, Deniss and Wu, Yuxiang},
  journal={arXiv preprint arXiv:2502.13138},
  year={2025}
}

@inproceedings{toledo2025airesearchagents,
  title={{AI} Research Agents for Machine Learning: Search, Exploration, and Generalization in {MLE}-bench},
  author={Toledo, Edan and Hambardzumyan, Karen and Josifoski, Martin and Hazra, Rishi and Baldwin, Nicolas and Audran-Reiss, Alexis and Kuchnik, Michael and Magka, Despoina and Jiang, Minqi and Lupidi, Alisia Maria and Lupu, Andrei and Raileanu, Roberta and Foerster, Jakob Nicolaus and Bachrach, Yoram},
  booktitle={Advances in Neural Information Processing Systems (NeurIPS)},
  year={2025}
}

@inproceedings{chan2025mlebench,
  title={{MLE}-bench: Evaluating Machine Learning Agents on Machine Learning Engineering},
  author={Chan, Jun Shern and Chowdhury, Neil and Jaffe, Oliver and Aung, James and Sherburn, Dane and Mays, Evan and Starace, Giulio and Liu, Kevin and Maksin, Leon and Patwardhan, Tejal and Weng, Lilian and Madry, Aleksander},
  booktitle={International Conference on Learning Representations (ICLR)},
  year={2025}
}

@article{wijk2024rebench,
  title={{RE}-Bench: Evaluating Frontier {AI} {R\&D} Capabilities of Language Model Agents against Human Experts},
  author={Wijk, Hjalmar and Lin, Tao and Becker, Joel and Jawhar, Sami and Parikh, Neev and Broadley, Thomas and Chan, Lawrence and Chen, Michael and Clymer, Josh and Dhyani, Jai and Ericheva, Elena and Garcia, Katharyn and Goodrich, Brian and Jurkovic, Nikola and Kinniment, Megan and Lajko, Aron and Nix, Seraphina and Sato, Lucas and Saunders, William and Taran, Maksym and West, Ben and Barnes, Elizabeth},
  journal={arXiv preprint arXiv:2411.15114},
  year={2024}
}

@article{nathani2025mlgym,
  title={{MLGym}: A New Framework and Benchmark for Advancing {AI} Research Agents},
  author={Nathani, Deepak and Madaan, Lovish and Roberts, Nicholas and Bashlykov, Nikolay and Menon, Ajay and Moens, Vincent and Budhiraja, Amar and Magka, Despoina and Vorotilov, Vladislav and Chaurasia, Gaurav and Hupkes, Dieuwke and Cabral, Ricardo Silveira and Shavrina, Tatiana and Foerster, Jakob and Bachrach, Yoram and Wang, William Yang and Raileanu, Roberta},
  journal={arXiv preprint arXiv:2502.14499},
  year={2025}
}

@article{grosnit2024agentk,
  title={Large Language Models Orchestrating Structured Reasoning Achieve Kaggle Grandmaster Level},
  author={Grosnit, Antoine and Maraval, Alexandre and Doran, James and Paolo, Giuseppe and Thomas, Albert and Beevi, Refinath Shahul Hameed Nabeezath and Gonzalez, Jonas and Khandelwal, Khyati and Iacobacci, Ignacio and Benechehab, Abdelhakim and Cherkaoui, Hamza and El-Hili, Youssef Attia and Shao, Kun and Hao, Jianye and Yao, Jun and K{\'e}gl, Bal{\'a}zs and Bou-Ammar, Haitham and Wang, Jun},
  journal={arXiv preprint arXiv:2411.03562},
  year={2024}
}

@inproceedings{yang2024sweagent,
  title={{SWE}-agent: Agent-Computer Interfaces Enable Automated Software Engineering},
  author={Yang, John and Jimenez, Carlos E. and Wettig, Alexander and Lieret, Kilian and Yao, Shunyu and Narasimhan, Karthik and Press, Ofir},
  booktitle={Advances in Neural Information Processing Systems (NeurIPS)},
  year={2024}
}

@inproceedings{devlin2019bert,
  title={{BERT}: Pre-training of Deep Bidirectional Transformers for Language Understanding},
  author={Devlin, Jacob and Chang, Ming-Wei and Lee, Kenton and Toutanova, Kristina},
  booktitle={Conference of the North American Chapter of the Association for Computational Linguistics (NAACL-HLT)},
  pages={4171--4186},
  year={2019}
}

@inproceedings{austin2021d3pm,
  title={Structured Denoising Diffusion Models in Discrete State-Spaces},
  author={Austin, Jacob and Johnson, Daniel D. and Ho, Jonathan and Tarlow, Daniel and van den Berg, Rianne},
  booktitle={Advances in Neural Information Processing Systems (NeurIPS)},
  year={2021}
}

@inproceedings{szegedy2016inception,
  title={Rethinking the Inception Architecture for Computer Vision},
  author={Szegedy, Christian and Vanhoucke, Vincent and Ioffe, Sergey and Shlens, Jon and Wojna, Zbigniew},
  booktitle={IEEE Conference on Computer Vision and Pattern Recognition (CVPR)},
  pages={2818--2826},
  year={2016}
}

@inproceedings{muller2019labelsmoothing,
  title={When Does Label Smoothing Help?},
  author={M{\"u}ller, Rafael and Kornblith, Simon and Hinton, Geoffrey},
  booktitle={Advances in Neural Information Processing Systems (NeurIPS)},
  year={2019}
}

@article{seong2026multimodal,
  title={Multimodal Crystal Flow: Any-to-Any Modality Generation for Unified Crystal Modeling},
  author={Seong, Kiyoung and Ahn, Sungsoo and Han, Sehui and Park, Changyoung},
  journal={arXiv preprint arXiv:2602.20210},
  year={2026}
}

@article{shazeer2020glu,
  title={GLU Variants Improve Transformer},
  author={Shazeer, Noam},
  journal={arXiv preprint arXiv:2002.05202},
  year={2020}
}

@inproceedings{xiong2020layer,
  title={On Layer Normalization in the Transformer Architecture},
  author={Xiong, Ruibin and Yang, Yunchang and He, Di and Zheng, Kai and Zheng, Shuxin and Xing, Chen and Zhang, Huishuai and Lan, Yanyan and Wang, Liwei and Liu, Tie-Yan},
  booktitle={Proceedings of the 37th International Conference on Machine Learning},
  pages={10524--10533},
  year={2020},
  volume={119},
  series={Proceedings of Machine Learning Research},
  publisher={PMLR}
}

@inproceedings{dehghani2023scaling,
  title={Scaling Vision Transformers to 22 Billion Parameters},
  author={Dehghani, Mostafa and Djolonga, Josip and Mustafa, Basil and Padlewski, Piotr and Heek, Jonathan and others},
  booktitle={International Conference on Machine Learning (ICML)},
  year={2023}
}

@misc{anthropic2026claudeopus47,
  title        = {Claude Opus 4.7},
  author       = {{Anthropic}},
  year         = {2026},
  month        = apr,
  howpublished = {\url{https://www.anthropic.com/news/claude-opus-4-7}},
  note         = {Accessed: 2026-06-14}
}

@article{liu2026autoresearchclaw,
  title={AutoResearchClaw: Self-Reinforcing Autonomous Research with Human-AI Collaboration},
  author={Liu, Jiaqi and Qiu, Shi and Li, Mairui and Li, Bingzhou and Ji, Haonian and Han, Siwei and Ye, Xinyu and Xia, Peng and Dong, Zihan and Zhang, Congyu and others},
  journal={arXiv preprint arXiv:2605.20025},
  year={2026}
}

@article{gao2026autoscientists,
  title={AutoScientists: Self-Organizing Agent Teams for Long-Running Scientific Experimentation},
  author={Gao, Shanghua and Fang, Ada and Zitnik, Marinka},
  journal={arXiv preprint arXiv:2605.28655},
  year={2026}
}

@misc{karpathy2026autoresearch,
  author       = {Karpathy, Andrej},
  title        = {{AutoResearch}: AI Agents Running Research on Single-GPU Nanochat Training Automatically},
  year         = {2026},
  howpublished = {\url{https://github.com/karpathy/autoresearch}},
  note         = {GitHub repository. Accessed: 2026-06-15}
}

@article{veljkovic2026crystalite,
  title={Crystalite: A Lightweight Transformer for Efficient Crystal Modeling},
  author={Veljkovi{\'c}, Tin Had{\v{z}}i and Rosenthal, Joshua and Lon{\v{c}}ari{\'c}, Ivor and van de Meent, Jan-Willem},
  journal={arXiv preprint arXiv:2604.02270},
  year={2026}
}

@inproceedings{kwon2023reward,
  title={Reward Design with Language Models},
  author={Kwon, Minae and Xie, Sang Michael and Bullard, Kalesha and Sadigh, Dorsa},
  booktitle={International Conference on Learning Representations (ICLR)},
  year={2023}
}

@article{zou2025humanagent,
  title={{LLM}-Based Human-Agent Collaboration and Interaction Systems: A Survey},
  author={Zou, Henry Peng and others},
  journal={arXiv preprint arXiv:2505.00753},
  year={2025}
}
\bibliographystyle{abbrvnat}

\clearpage
\appendix
\crefalias{section}{appendix}
\section{Algorithms}
\begin{algorithm}[h]
\caption{MaskGXT training}
\label{alg:training}
\begin{algorithmic}[1]
\STATE \textbf{Input:} dataset $\mathcal{D}$, network $f_\theta$, bins $K$, stream weights $\{w_m\}$, smoothing $\sigma$.
\STATE Precompute the circular ordinal soft-target matrix $p^{\star}$ (\cref{eq:softtarget}).
\WHILE{not converged}
  \STATE Sample a crystal $(\mathbf{L},\mathbf{F},\mathbf{A},g,\mathbf{W}_g)\sim\mathcal{D}$.
  \STATE Draw an origin shift $\mathbf{r}\sim\mathcal{U}[0,1)^3$, translate all coordinates by $\mathbf{r}$ (mod $1$), and relabel Wyckoff letters by the normalizer coset nearest to $\mathbf{r}$.
  \STATE Order the sites canonically by symmetry orbit (electronegativity, then Wyckoff letter), and apply the intra/inter-orbit permutation augmentation.
  \STATE Quantize $\mathbf{L},\mathbf{F}$ into $K$ bins (\cref{eq:quantize}) and assemble the token streams $\mathcal{C}$.
  \STATE Draw $t\sim\mathcal{U}(0,1)$ and mask each token independently with probability $t$, giving $\mathcal{C}_{\setminus\mathcal{M}}$.
  \STATE Predict token logits $f_\theta(\mathcal{C}_{\setminus\mathcal{M}})$ at the masked positions $\mathcal{M}$.
  \STATE Compute the masked cross-entropy $\mathcal{L}(\theta)$ against the soft targets (\cref{eq:loss}).
  \STATE Update $\theta$ with AdamW.
\ENDWHILE
\end{algorithmic}
\vspace{-.05in}
\end{algorithm}
\begin{algorithm}[h]
\caption{MaskGXT sampling}
\label{alg:sampling}
\begin{algorithmic}[1]
\STATE \textbf{Input:} composition $\mathbf{A}$, steps $T$, network $f_\theta$; optional fixed space group $g^{\dagger}$.
\STATE Initialize all lattice, coordinate, space group, and Wyckoff tokens to \textsc{mask}; if a space group $g^{\dagger}$ is given, fix its token to $g^{\dagger}$ throughout decoding.
\FOR{$\tau = 1, \dots, T$}
  \STATE Predict the posterior $\softmax(f_\theta)$ for all tokens in parallel.
  \STATE At each masked position, draw a token $\hat{c}\sim\softmax(f_\theta)$. \\
  $\triangleright$ greedy decoding: $\hat{c}=\argmax \softmax(f_\theta)$
  \STATE Score each position by the posterior probability (confidence) of its drawn token $\hat{c}$.
  \STATE Fix the drawn tokens at the highest-confidence positions and re-mask the rest.
\ENDFOR
\STATE Dequantize lattice and coordinate refined by sub-bin offset prediction and \textbf{return} $(\mathbf{L},\mathbf{F})$.
\end{algorithmic}
\end{algorithm}
\begin{algorithm}[h]
\caption{Polymorph-aware sampling}
\label{alg:polymorph}
\begin{algorithmic}[1]
\STATE \textbf{Input:} composition $\mathbf{A}$, stratification count $S$, steps $T$, network $f_\theta$.
\STATE Read the space group posterior $p(g)=\softmax\!\big(f_\theta(\mathcal{C}_{\mathrm{mask}})_{\mathrm{sg}}\big)$ from one forward pass on the fully masked sequence $\mathcal{C}_{\mathrm{mask}}$.
\STATE Take the $S$ highest-posterior space groups $\mathcal{G}=\operatorname{top\text{-}}S_g\, p(g)$.
\STATE For each $g\in\mathcal{G}$, decode $(\mathbf{L}_g,\mathbf{F}_g)$ by MaskGXT sampling (\cref{alg:sampling}) with the space group fixed to $g$.
\STATE \textbf{return} $\{(\mathbf{L}_g,\mathbf{F}_g)\}_{g\in\mathcal{G}}$.
\end{algorithmic}
\end{algorithm}

\newpage
\section{Hyperparameters}
\label{app:config}
\begin{table}[h]
\centering
\caption{MaskGXT configuration.}
\label{tab:config}
\begin{tabular}{ll}
\toprule
Component & Value \\
\midrule
Coordinate / lattice bins $K$ & $64$ \\
Transformer layers & $34$ \\
Hidden dimension & $768$ \\
Attention heads & $12$ (QK-normalized) \\
Feed-forward & SwiGLU \\
Parameters & $\approx 248$M \\
Ordinal label-smoothing $\sigma$ & $1.0$ (coordinate and lattice) \\
Stream weights $(w_{\mathrm{lat}}, w_{\mathrm{coord}}, w_{\mathrm{sg}}, w_{\mathrm{wyk}})$ & $(0.10,\,1.5,\,0.10,\,0.30)$ \\
Offset-regression loss weight (lattice, coordinate) & $(0.2,\,0.2)$ \\
Euclidean-normalizer augmentation probability & $0.7$ \\
Permutation augmentation probability & $0.5$ \\
Dropout & $0.05$ \\
Optimizer & AdamW ($\beta_1{=}0.9$, $\beta_2{=}0.95$, weight decay $0.05$) \\
Learning rate & $4\times10^{-4}$, cosine decay to $5\%$ floor \\
Batch size & $512$ ($128$ for MPTS-52) \\
Gradient clipping & $1.0$ \\
Weight EMA decay & $0.9999$ \\
Training epochs & up to $3000$ with early stopping on validation METRe \\
Sampling steps $T$ & $150$ \\

\bottomrule
\end{tabular}
\end{table}

\paragraph{Low-confidence fallback in stratified sampling.}
Space group stratified sampling (\cref{alg:polymorph}) clamps a branch to a space group only when that space group holds at least $0.02$ of the first-step posterior mass. When every space group is below this threshold, the branch leaves the space group free and draws the space group and remaining tokens i.i.d.\ rather than greedily.

\newpage
\section{Uniform Sampling Budget Sweep}
\label{app:sweep}

\begin{figure}[t]
\centering
\includegraphics[width=\linewidth]{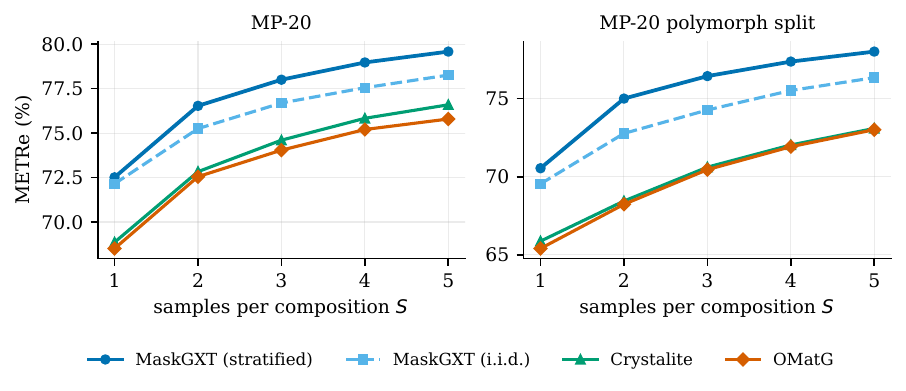}
\caption{
\textbf{METRe versus the uniform sampling budget $S$.}
Each composition receives $S$ structures and METRe is measured against the full
reference set, on MP-20 (left) and the MP-20 polymorph split (right). MaskGXT
with space group stratified sampling has the highest METRe at every $S$ on both
splits.
}
\label{fig:sweep}
\end{figure}

\Cref{tab:uniform} fixes the per-composition budget to $S=2$. \Cref{fig:sweep} sweeps $S$ from one to five under the same uniform-sampling protocol, where every composition receives the same budget and METRe is measured against the full reference set. MaskGXT with space group stratified sampling leads on both splits across the entire range. The baselines narrow the gap only as $S$ increases: additional uniform draws yield the largest gains when the single-structure accuracy ($S=1$) is low, so the weaker baselines climb faster while MaskGXT, already accurate at $S=1$, improves more gradually.

\newpage
\section{Details of AI Co-Scientist Run}
\label{app:dialogue}
We reproduce three turning points from the search verbatim, styled as a transcript.
Teal boxes are autonomous \textbf{Agent} turns, labeled with the search operator
that produced them (\emph{idea}, \emph{draft}, \emph{improve}); orange boxes are
sparse high-level \textbf{Human} interventions; gray boxes are the
\textbf{Orchestrator}, the sole interface for human steering, which routes each
intervention into the operator it dispatches. Nodes are shown in the order they
occurred along the winning lineage. Human interventions take two forms: an
\emph{objective} intervention introduces a new target the agent was not yet
pursuing and leaves the method open, whereas a \emph{mechanism} intervention
supplies a causal lever toward the existing objective and lets the agent
operationalize it.

\medskip
\noindent\textbf{(a) Autonomous cross-domain transfer.}\hfill\nodetag{Node 18}
\par\smallskip
\begin{agentturn}
\agenthead{idea}
Survey generative frameworks outside materials science and filter them against a catalogue of published CSP methods. \emph{Generative framework choice}: masked generative modeling, the framework underlying MaskGIT in vision. Unlike diffusion and flow baselines, which operate on the continuous domain of the fractional-coordinate torus, discrete modeling with quantized tokens supports categorical posteriors, uses standard transformer architectures, and decodes in a few parallel unmasking steps rather than the hundreds of denoising steps required by those baselines. \emph{Architecture choice}: a bidirectional transformer, since predicting a masked token requires attending to all visible tokens simultaneously, and a crystal's variable-length, set-like atom sequence is well suited to attention over a fixed grid. Novelty: this combination is absent from the published-CSP catalogue. Feasibility: a crystal's periodicity can be encoded with discrete tokens.
\end{agentturn}
\begin{agentturn}
\agenthead{draft}
Instantiating the cell in code. I discretize each fractional coordinate axis into $K$ bins (a resolution within the structure-matching tolerance) to form coordinate tokens, and Niggli-parameterize the unit cell into six scalars ($a,b,c,\alpha,\beta,\gamma$), each binned over data-derived ranges, as lattice tokens; six scalars are a cleaner target than the nine-entry lattice matrix. The composition is given, so I embed the atomic numbers as fixed per-site side information rather than generating them, leaving six lattice tokens plus $3N$ coordinate tokens to predict. I keep equivariance implicit in the representation: fractional coordinates are rotation-invariant and the lattice tokens carry only invariant Niggli scalars, so a plain bidirectional transformer over the tokens suffices without vector features. I train it with absorbing-state masking and a cross-entropy loss on the masked positions, sample by iterative parallel unmasking from MaskGIT, and train against the validation METRe.
\end{agentturn}

\medskip
\noindent\textbf{(b) Objective intervention: recover sub-bin coordinate precision.}\hfill\nodetag{Node 340}
\par\smallskip
\begin{humanturn}
\humanhead
Quantizing coordinates into bins caps geometric precision. Recover the lost precision without leaving the discrete token model.
\end{humanturn}
\begin{orchturn}
\orchhead
I dispatch the \emph{improve} operator on the current best node, folding this objective into its prompt.
\end{orchturn}
\begin{agentturn}
\agenthead{improve}
I add a small regression head that reads the trunk output at each coordinate and lattice position and predicts a bounded sub-bin offset, supervised against the signed residual about the bin center and kept out of the masked cross-entropy objective. At decode time, I place each value at its bin center shifted by the predicted offset, so precision is no longer floored at the bin width.
\end{agentturn}

\medskip
\noindent\textbf{(c) Mechanism intervention: cover multiple polymorphs.}\hfill\nodetag{Node 391}
\par\smallskip
\begin{humanturn}
\humanhead
i.i.d.\ sampling draws independently, so a composition's distinct polymorphs go uncovered. Make the sampler spread its draws across those polymorphs.
\end{humanturn}
\begin{orchturn}
\orchhead
I dispatch the \emph{improve} operator in the sampling-optimization stage, folding this mechanism into its prompt.
\end{orchturn}
\begin{agentturn}
\agenthead{improve}
Since the space group is an explicit token, I read its posterior from one forward pass on the fully masked sequence, conditioned only on the composition. Within each composition I pin the most confident draw to its argmax space group and assign the remaining draws to distinct high-posterior space groups without replacement. The diversity now comes from space group conditioning rather than sampling noise.
\end{agentturn}

\end{document}